\documentclass[10pt,twocolumn,letterpaper]{article}

\usepackage{cvpr}              

\usepackage{graphicx}
\usepackage{amsmath}
\usepackage{amssymb}
\usepackage{booktabs}
\usepackage{float}
\usepackage{multirow}
\usepackage{bm}

\usepackage[pagebackref,breaklinks,colorlinks]{hyperref}

\usepackage[capitalize]{cleveref}
\crefname{section}{Sec.}{Secs.}
\Crefname{section}{Section}{Sections}
\Crefname{table}{Table}{Tables}
\crefname{table}{Tab.}{Tabs.}

\def\cvprPaperID{8042} 
\def\confName{CVPR}
\def\confYear{2022}

\newcommand\blfootnote[1]{%
  \begingroup
  \renewcommand\thefootnote{}\footnote{#1}%
  \addtocounter{footnote}{-1}%
  \endgroup
}

\begin{document}

\title{GCFSR: a Generative and Controllable Face Super Resolution Method \\
Without Facial and GAN Priors}

\author{Jingwen He$^{1}$ \quad Wu Shi$^{2}$ \quad Kai Chen$^{1}$ \quad Lean Fu$^{1}$ \quad Chao Dong$^{2,3\, *}$\\
${}^{1}$ByteDance Inc\\ ${}^{2}$Shenzhen Institute of Advanced Technology, Chinese Academy of Sciences\\
${}^{3}$Shanghai AI Laboratory, Shanghai, China
}

\twocolumn[{%
\maketitle
\begin{figure}[H]
\hsize=\textwidth 
\centering
\includegraphics[scale=0.18]{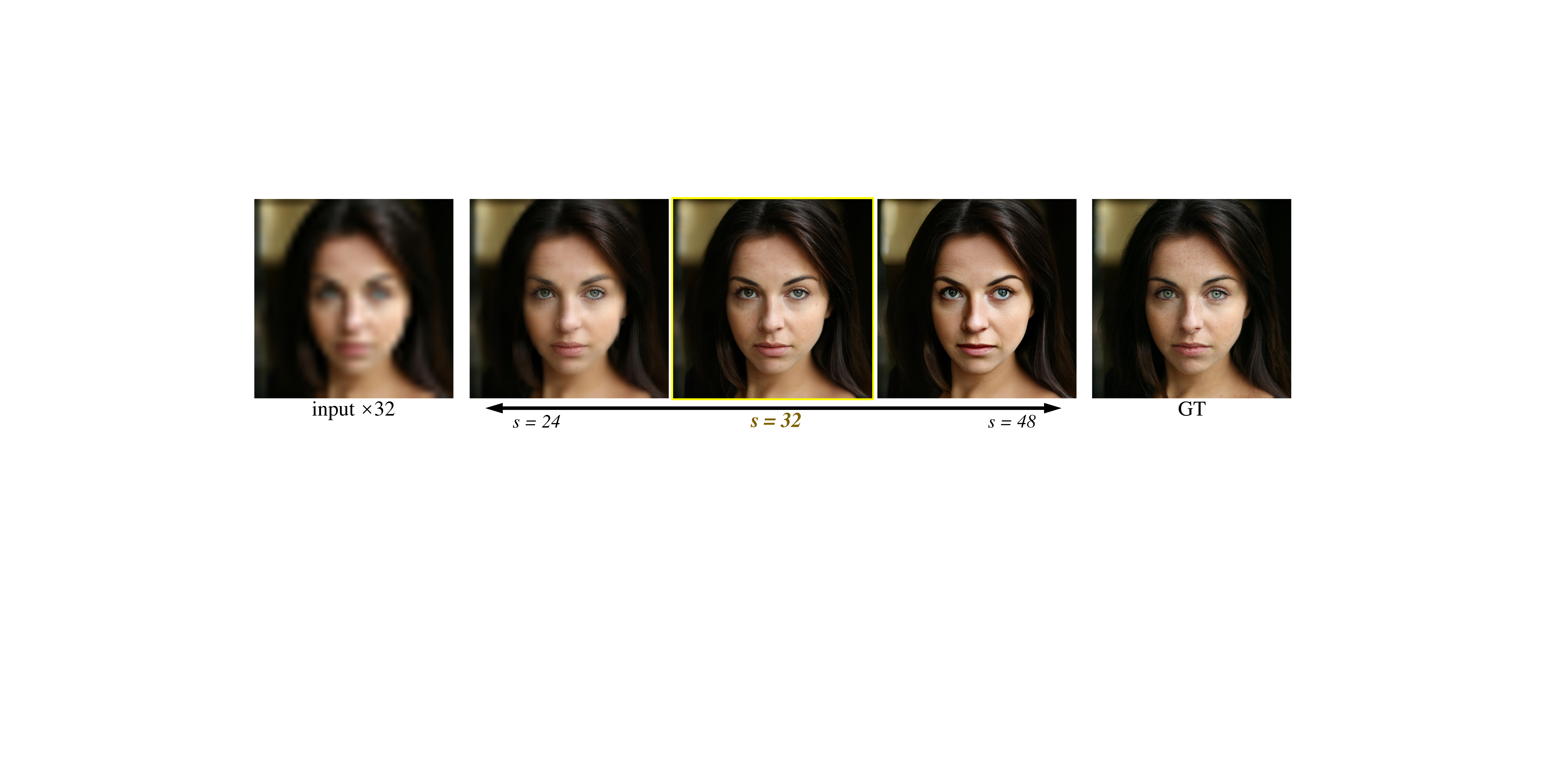}
\caption{The results of our proposed GCFSR on CelebA-HQ \cite{celeb} for $32\times$ SR.  Left: The original $32\times32$ LR input image is upsampled ($\times32$) to a resolution $1024^2$ by bicubic interpolation. The middle three images are the modulated results of GCFSR under different conditional upscaling factors ($s=24, s=32, s=48$). The conditional upscaling factor $s$ controls the generative strength. The best SR result ($s=32$) is denoted by a yellow rectangle.
Right: The ground truth (GT) at $1024^2$ resolution.
\textbf{(Zoom in for best view)}}
\label{fig:figure1}
\end{figure}
}]

\begin{abstract}\blfootnote{* Corresponding author (e-mail: chao.dong@siat.ac.cn)}
Face image super resolution (face hallucination) usually relies on facial priors to restore realistic details and preserve identity information. 
Recent advances can achieve impressive results with the help of GAN prior. 
They either design complicated modules to modify the fixed GAN prior or adopt complex training strategies to finetune the generator. 
In this work, we propose a generative and controllable face SR framework, called GCFSR, which can reconstruct images with faithful identity information without any additional priors. 
Generally, GCFSR has an encoder-generator architecture. 
Two modules called style modulation and feature modulation are designed for the multi-factor SR task. 
The style modulation aims to generate realistic face details and the feature modulation dynamically fuses the multi-level encoded
features and the generated ones conditioned on the upscaling
factor. 
The simple and elegant architecture can be trained from scratch in an end-to-end manner. 
For small upscaling factors ($\leq$8), GCFSR can produce surprisingly good results with only adversarial loss. 
After adding L1 and perceptual losses, GCFSR can outperform state-of-the-art methods for large upscaling factors (16, 32, 64). 
During the test phase, we can modulate the generative strength via feature modulation by changing the conditional upscaling factor continuously to achieve various generative effects. 
\end{abstract}

\section{Introduction}
\label{sec:intro}

Face image super resolution (face SR or face hallucination) algorithms have been developing rapidly in recent years, with its wide application in video restoration and AI photographing. Face SR has a close relationship with general image SR \cite{srcnn, fsrcnn, esrgan, sftgan, ranksrgan, bsrgan, realesrgan} and face generation tasks \cite{stylegan1, stylegan2, progan, stylegan3, swagan}. Similar as general image SR, face SR is a restoration problem, whose goal is to reconstruct correct structures and preserve identity information. Differently, face SR has to deal with very large upscaling factors (8-64) \cite{gpen, gfpgan, glean}, thus requiring to generate a large amount of finer details, which is similar 
to face generation. As a combination of restoration and generation problem, face SR has unique solution pipelines, which always involve various additional facial priors \cite{fsrnet, bulat2018super, kim2019progressive, zhu2016deep, kim2019progressive, yu2018face, shen2018deep, psfrgan}, like parsing map and attribution map. 

Recent advances have found that a face GAN can take the place of all previous facial priors, and produce realistic face details. This is based on the observation that a well-trained GAN model has already contained enough face information, which is sufficient to provide strong priors. For instance, GLEAN \cite{glean} adopts the intermediate features of a StyleGAN \cite{stylegan2} as latent banks, and achieves superior performance on large-factor SR tasks. While GFPGAN \cite{gfpgan} and GPEN \cite{gpen} introduce face GAN models to solve blind face restoration problem, and both can recover promising facial details. Their success can be attributed to the utilization of GAN priors and motivates later works to find more applications.

However, if we must rely on such a prior, face SR will face two apparent limitations. First, as face GAN is trained on specific datasets (e.g., FFHQ \cite{ffhq}), the corresponding face SR methods can only deal with the same kind of face images (e.g., frontal faces with a fixed size), significantly restricting its applications. Second, as face GAN is not specially designed for face SR, we have to add additional operations in the network for adaptation \cite{gpen, gfpgan, glean, dgp}, which is a waste of computation resources. Then we will ask: can we get rid of these priors, and design a pure data-driven framework? 

Another issue unsolved in face SR is the flexibility of generation. Existing methods can only output a single restoration result with a fixed style. However, in real scenarios, users might want to adjust the generative strength to meet personalized requirements. For example, they will desire more details in old photo restoration, but less hallucination effects in surveillance video enhancement. ``How to control the amount of generated details" is a practical demand. Furthermore, real-world images may have various sizes, but conventional SR models (e.g., GLEAN \cite{glean} and ESRGAN \cite{esrgan}) for fixed upscaling factors cannot handle such diverse cases. 

To address the problems, we propose a generative and controllable face SR framework, called GCFSR, which has three appealing properties. First, it could reconstruct faithful images with promising identity information. This is also the basic requirement of face SR task. Second, it could generate realistic face details, without reliance on any additional priors, including facial priors and GAN priors. This shows that GAN prior is not an essential part in face SR task. Third, its generative strength can be interactively adjusted (Figure \ref{fig:figure1}). This can also be used in handling different and continuous upscaling factors. These three properties are guaranteed by three special designs in GCFSR, which are the encoder-generator architecture, style modulation and feature modulation modules. GCFSR enjoys a very concise architecture without extra priors or initialization. We will detail our designs in the Method section. More importantly, GCFSR has a nice training property. It is end-to-end trainable and converges fast. When the upscaling factor is small ($\leq8$), it is possible to discard all pixel-wise constraints and use a single GAN loss to achieve state-of-the-art performance. This has never been revealed in previous SR methods. Extensive experiments and ablation studies have demonstrated the effectiveness of each module. Combing them together, GCFSR could achieve superior performance to GAN based methods in both small and large upscaling factors. We can also observe vivid face details and gradually modulated effects in qualitative results (see Figure \ref{fig:figure1}).

\section{Related Work}

\subsection{Face Super Resolution.}
We can divide face super resolution works into two groups according to the utilization of facial priors. On the one hand, the works in \cite{yu2016ultra, zhang2018super, cao2017attention, tuzel2016global, huang2017wavelet} directly use deep neural networks for face SR without any additional facial priors.
On the other hand, more recent works focus on the investigation in facial priors to preserve the identity information as well as generate faithful face details. In general, facial landmarks \cite{fsrnet, bulat2018super, kim2019progressive, zhu2016deep, kim2019progressive}, face parsing maps \cite{shen2018deep, psfrgan} and facial attributes \cite{yu2018face} have been demonstrated to be effective in the face image reconstruction. Chen et al. \cite{fsrnet} predict landmark heatmaps and parsing maps from LR faces, then use them to further finetune the SR results. 
Differently, the work in \cite{bulat2018super} learns face SR and landmark prediction jointly. 
Yu et al. \cite{yu2018face} utilize a convolution neural network to obtain face component heatmaps in order to achieve improvement for face super resolution.

Recently, significant advances have been made by using face GAN prior \cite{glean, gfpgan, gpen} instead of the previous facial priors. Generally, these state-of-the-art methods all design additional modules to extract feature maps and latent vectors, then use them to adapt the face GAN prior to handle face super resolution/restoration tasks.

\subsection{GAN Prior}
The pretrained GAN priors \cite{stylegan1, stylegan2, biggan} have been deeply exploited in GAN inversion \cite{zhujiapeng_inversion, pulse, mganprior, dgp, psp}. In PULSE \cite{pulse}, the latent code of GAN prior is iteratively optimized with L1 constraint between the input and downsampled output. While mGANprior \cite{mganprior} optimizes multiple latent codes to improve the capacity for reconstruction. Instead of relying on latent code alone, DGP \cite{dgp} also finetunes the pretrained GAN prior for better performance. Since the GAN inversion methods only use the insufficient low-dimension latent code for image reconstruction, they usually generate undesirable results with low fidelity. To solve this issue, GLEAN \cite{glean} uses an additional RRDBNet \cite{esrgan} to extract the multi-resolution features, which will be fused with the intermediate features in GAN prior. GLEAN is designed for SR on single upscaling factor and achieves the state-of-the-art performance. GPEN \cite{gpen} and GFPGAN \cite{gfpgan} both achieve the state-of-the-art for blind face restoration (BFR) problem.
The two GAN-prior-based methods also adopt additional encoders to extract multi-resolution features and combine them with intermediate features of the pretrained GAN prior. 
More descriptions of these state-of-the-art GAN-prior-based methods could be found in Method \ref{sec:method_rethinking}.

Different from them, our proposed GCFSR adopts an end-to-end training strategy without any additional priors (facial prior or GAN prior), and achieves the state-of-the-art performance on face SR. In addition, our method provides the flexibility for user adjustment on the generative strength.

\section{Method}
\subsection{Rethinking on GAN-prior-based Methods.}
\label{sec:method_rethinking}
Before introducing the proposed GCFSR, we give a brief review on previous state-of-the-art GAN-prior-based image restoration/super resolution methods: GLEAN \cite{glean}, GFPGAN \cite{gfpgan}, and GPEN \cite{gpen}. Here we provide the detailed descriptions of these methods in Table \ref{table:observation}.

\begin{table}[htbp]
\centering
\small
\caption{The detailed descriptions of the state-of-the-art GAN-prior-based methods: GLEAN \cite{glean}, GFPGAN \cite{gfpgan}, GPEN \cite{gpen}.}
\vspace{-1em}
\label{table:observation}
\begin{tabular}{|l|c|c|c|}
\hline
 & \textbf{GLEAN} \cite{glean} & \textbf{GFPGAN} \cite{gfpgan} & \textbf{GPEN} \cite{gpen} \\ \hline\hline
Degradation & single & multiple & multiple \\ \hline
\begin{tabular}[c]{@{}l@{}}Network\\ Description\end{tabular}  & \begin{tabular}[c]{@{}l@{}}encoder\\GAN prior\\decoder\end{tabular} & \begin{tabular}[c]{@{}l@{}}encoder\\CS-SFT\\GAN prior\end{tabular} & \begin{tabular}[c]{@{}l@{}}encoder\\Concat\\GAN prior\end{tabular} \\ \hline
\begin{tabular}[c]{@{}l@{}}Parameters\end{tabular} & $188.29$M  & $90.76$M & $71.21$M \\ \hline
GAN prior & fixed & fixed & finetuned \\ \hline
\end{tabular}

\end{table}

\begin{figure}[htbp]
	\centering
	\includegraphics[scale=0.43]{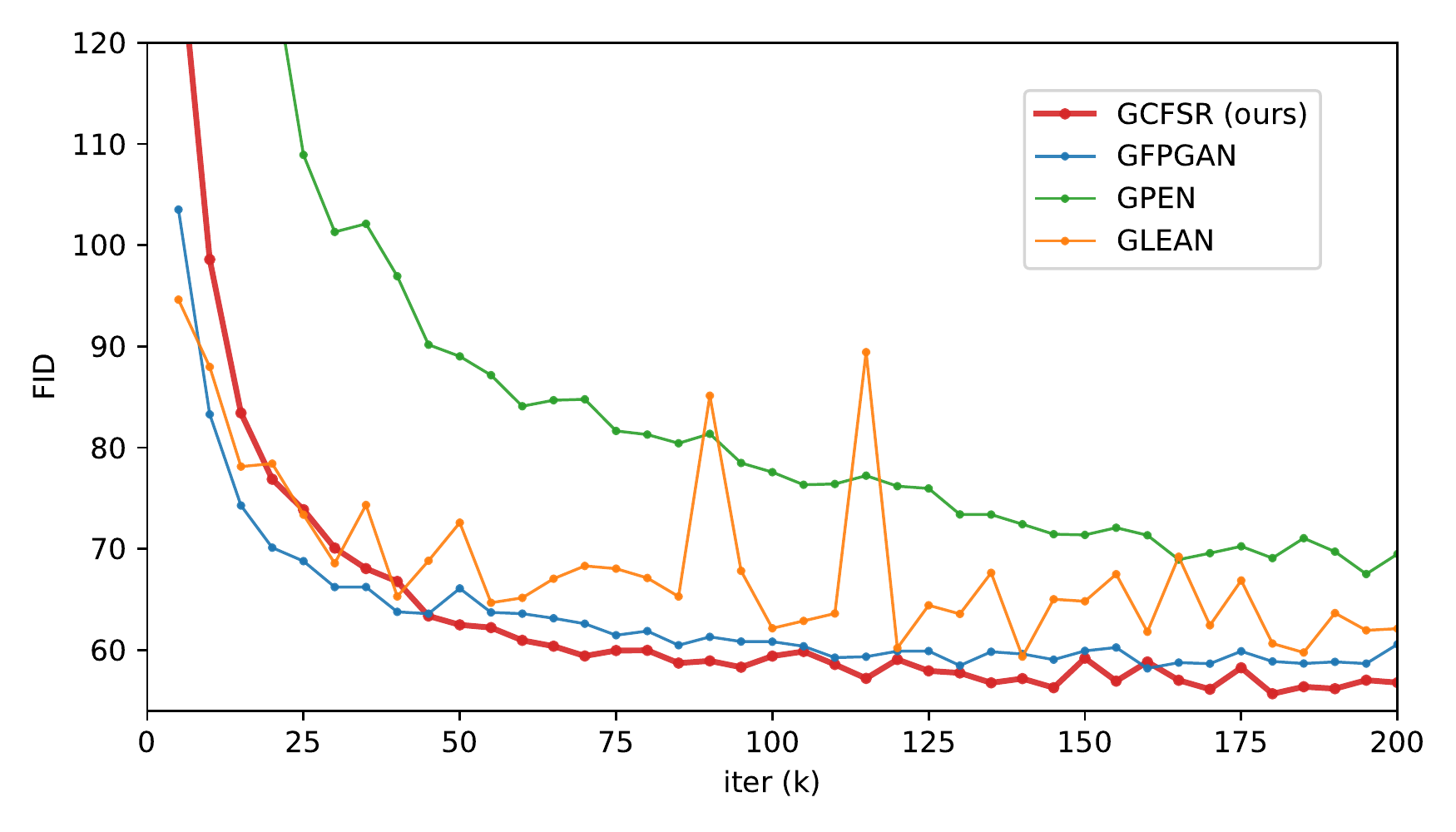}
	\vspace{-1em}
	\caption{The convergence curves of GCFSR (ours), GFPGAN \cite{gfpgan}, GPEN \cite{gpen}, and GLEAN \cite{glean}. The x and y axes denote the training iterations (k) and FID scores on CelebA-HQ for $64\times$ SR.}
	\vspace{-1em}
	\label{fig:convergence}
\end{figure}

First of all, GLEAN \cite{glean} is proposed for image super resolution on single upscaling factor, while GFPGAN and GPEN could deal with multiple degradations. 
From Table \ref{table:observation}, we observe that GLEAN relies heavily on additional modules for SR. Specifically, GLEAN utilizes a RRDBNet \cite{esrgan} for feature extraction, and then combines the fixed GAN prior with an additional decoder to generate the final output. Thus, GLEAN has significantly more parameters than the other two methods (Table \ref{table:observation}). 
Similarly, GFPGAN \cite{gfpgan} adopts an additional UNet \cite{unet} trained with L1 loss for the degradation removal, and then transforms the features in Unet to the parameters of scaling and shifting operations, which will be used to modify the fixed GAN prior. 
With the pretrained GAN prior, GLEAN and GFPGAN could achieve better performance than others at the beginning of training (Figure \ref{fig:convergence}).
Differently, GPEN \cite{gpen} directly concatenates the features from the encoder and GAN prior. As the concatenation operation introduces new parameters to GAN prior, GPEN gives the GAN prior a small learning rate for further finetuning.
However, as shown in Figure \ref{fig:convergence}, this strategy leads to slow convergence and inferior performance compared with other methods.

In conclusion, previous GAN-prior-based methods either design complicated modules to modify the fixed GAN prior, or further finetune the GAN prior for adaptation.
These observations indicate that utilizing GAN prior in face restoration/super resolution is not a trivial task. If this is the case, can we design a new generative model without reliance on the pretrained GAN prior? Thus, in this work, we propose a very concise architecture -- GCFSR without extra priors. As we can see from Figure \ref{fig:convergence}, the end-to-end trainable GCFSR converges fast and outperforms the state-of-the-art GAN-prior-based methods. 

\begin{figure*}[t]
\begin{center}
 \vspace{-1em}  
   \includegraphics[width=0.95\linewidth]{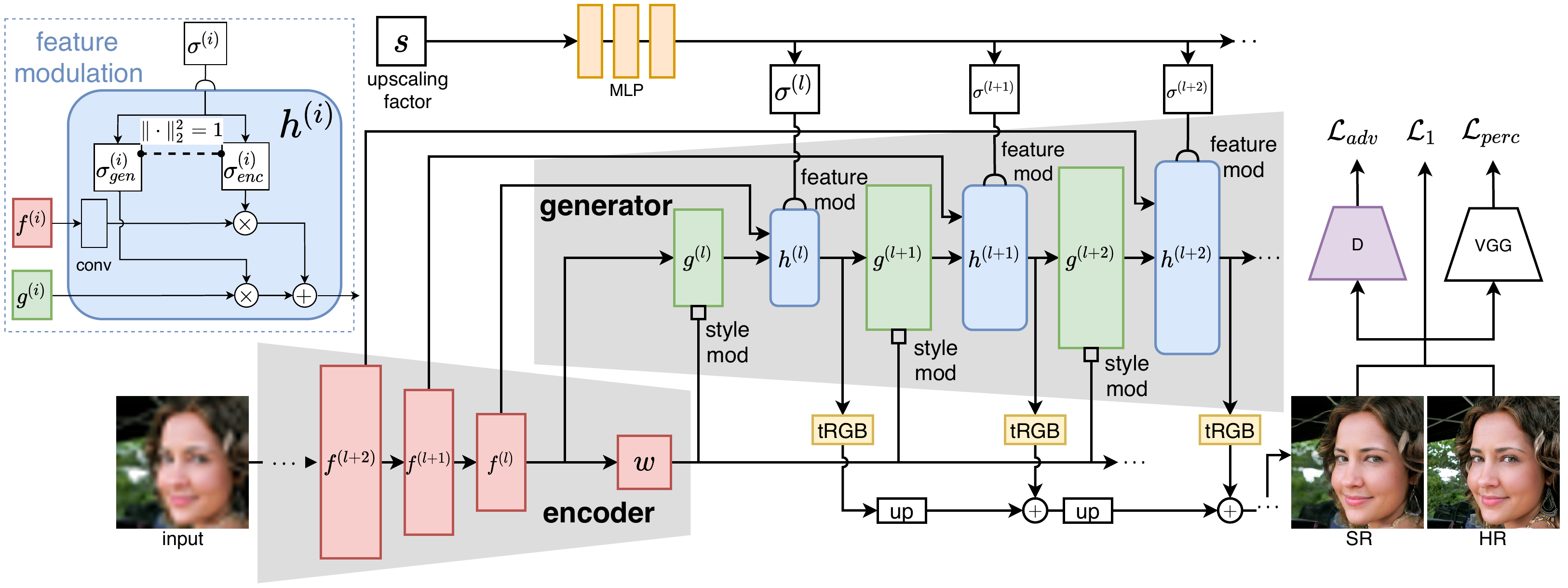}
\end{center}
\vspace{-2em}
   \caption{The architecture of GCFSR. It contains an encoder (red) and a generator (green \& blue).
   The encoder network uses several strided convolutional layers to extract the multi-level features and latent codes $\bm{w}$.
   The generator takes the topmost encoded feature maps and latent codes $\bm{w}$ to generate realistic face details by a sequence of style-modulated convolutions \cite{stylegan2}, namely as \textit{style modulation} (green) here. While the \textit{feature modulation} (blue) module controls how much the encoded and generated features are expressed under the conditional upscaling factor $s$. We train the whole network in an end-to-end manner.
   (The colored blocks are trained from scratch, while the other blocks are fixed or contain no trainable parameters.)}
 \vspace{-1em}  
\label{fig:overview}
\end{figure*}

\subsection{Overview of GCFSR}
Given an input LR face image $\bm{x} = \mathop{\downarrow_s}(\bm{y})$ and the upscaling factor $s$, GCFSR aims to estimate an SR face image $\hat{\bm{y}}$, which is as close as possible to its ground truth $\bm{y}$. 
To achieve this goal, GCFSR learns a mapping function $G(\bm{x}, s) \rightarrow \bm{y}$, where $s$ comes from a set of upscaling factors (e.g., $s\in\{4, 8, 16, 32, 64\}$), and the size of $\bm{y}$ is $2^{u}$ (e.g., 512, 1024). 
On the other hand, the target upscaling factor $s$ corresponds to the strength to generate the missing details in the down-scaling process. During testing, the generative strength could be smoothly adjusted by changing the conditional upscaling factor $s$ continuously.

The overall framework of GCFSR is depicted in Figure~\ref{fig:overview}. In general, it consists of an encoder network and a generator network. 
The encoder network takes the LR face image $\bm{x}$ as input and extracts the face structure roughly. It also estimates the latent codes $\bm{w}$ in $\mathcal{W}+$ space~\cite{stylegan2} for the generative process. 
The generator network takes the topmost encoded feature maps as well as the latent codes $\bm{w}$ to generate realistic face details by a sequence of style-modulated convolutions~\cite{stylegan2}.
To handle a wide range of upscaling factors, we add skip connections from the extracted structural features to the corresponding generated ones. Specifically, our proposed feature modulation module at each level controls how much the encoded and generated features are expressed under the conditional upscaling factor $s$. 
The two networks collaborate closely with each other, yielding realistic results with high fidelity and providing the flexibility for user adjustment.

The framework can be trained in an elegant end-to-end manner without any GAN prior pretraining or finetuning using complex learning objectives. 
Our method can obtain visually pleasing results using only adversarial loss for small upscaling factors ($4, 8$). After adding L1 and perceptual losses, our method can achieve state-of-the-art performance for large upscaling factors ($16, 32, 64$).


\subsection{Encoder Network}
The encoder network is a simple convolution neural network (CNN) with stride of 2. 
The intermediate features are denoted by $\{\dots, f^{(l+1)}, f^{(l)}\}$, where the superscript indicates the base 2 logarithm of feature size. The topmost feature map, $f^{(l)}$, has a size of $2^l$. The input LR image is resized by bicubic interpolation to the size of $2^u$, \ie, $\mathop{\uparrow_*}(\bm{x})\in \mathbb{R}^{2^u\times 2^u\times 3}$. Formally, we define
\begin{equation}
  f^{(i)} = 
  \begin{cases}
    \mathop{\mathrm{Conv}}(\mathop{\uparrow_*}(\bm{x})), & i = u,\\
    \mathop{\mathrm{Conv}}(f^{(i+1)}), & l\leq  i < u,
  \end{cases}
  \label{eqn:enc_feature}
\end{equation}
where $\mathop{\mathrm{Conv}}$ denotes a convolution layer with bias and activation.
The encoded features are used to carry multi-level structural information of the input image. Besides, we add several convolution layers and a fully-connected (FC) layer on the topmost feature to estimate the latent codes, $\bm{w} = [w^{(l)}, w^{(l+1)}_{1,2}, \dots]$, in $\mathcal{W}+$ space. The latent codes are further used by a style-based generator to generate realistic facial details. 
\begin{equation}
  \bm{w} = \mathop{\mathrm{Reshape}}(\mathop{\mathrm{FC}}(\mathop{\mathrm{Conv}}\cdots(f^{(l)}))).
  \label{eqn:enc_latent}
\end{equation}

\subsection{Generator Network}
%
%
The generator takes the topmost encoded feature maps and latent codes, $\bm{w}$, to generate realistic facial details by \emph{style modulation}. While the \emph{feature modulation} controls how much the encoded and generated features are expressed at each level given the conditional upscaling factor $s$.
We will elaborate on the details of the two modules as follows.
%
\paragraph{Style modulation.} 
Style-modulated convolution is proposed in StyleGAN2~\cite{stylegan2}. It uses a latent vector to modulate the convolution kernel on the input channel dimension. To approximately preserve the variances between input and output neurons, the kernel is  channel-wisely normalized before conducting the convolution. Bias, noise and activation are added to the output. (Please refer to the original paper for more details.) We denote the whole module by $\mathop{\mathrm{Conv_{sm}}}$. 
The generation starts from a 4-by-4 constant feature map, $c$. The feature map, $g^{(i)}$, progressively grows in size via up-sampling, $\mathop{\uparrow_2}$, and $\mathop{\mathrm{Conv_{sm}}}$. 

%
In our method, we make several modifications for adaptation to SR tasks.
First, we do not re-generate from the small constant feature map. Instead, we replace it by the topmost encoded feature, \ie, $c = f^{(l)}$ and let $l$ be equal to the minimum input size.
Then, our proposed feature modulation will join the multi-level encoded features ($\{\dots, f^{(l+1)}, f^{(l)}\}$) and the generated features ($\{\dots, g^{(l+1)}, g^{(l)}\}$) under the condition of upscaling factor $s$. The fused result is denoted by $h^{(i)}$ at level $i$.
Thus, the generator can benefit from both encoded and generated features and does not need to generate everything from scratch.
Formally, we define the style modulation as
\begin{equation}
  g^{(i)} = 
  \begin{cases}
    \mathop{\mathrm{Conv_{sm}}}(c, w^{(l)}), & i = l, \\
    \mathop{\mathrm{Conv_{sm}}}(\mathop{\mathrm{Conv_{sm}}}(\mathop{\uparrow_2}(h^{(i-1)}), w^{(i)}_1), w^{(i)}_2), & i > l.
   \end{cases}
   \label{eqn:style_mod}
\end{equation}

\paragraph{Feature modulation.} 
To handle the multi-factor SR in a single model, the amounts of input information to preserve and output details to generate may vary depending on the upscaling factor. 
%
It is difficult for the add/concat skip connection to fulfill the various requirements of consistency and generative capacity for different SR tasks.
We thus propose the feature modulation to flexibly adjust the generative strength. 
The upscaling factor, $s$, is first transformed by an MLP to a set of scaling vectors, $\bm{\sigma} = \{\sigma^{(l)}_{1,2}, \sigma^{(l+1)}_{1,2}, \dots\}$. At level $i$, $\sigma^{(i)}_{1}$ and $\sigma^{(i)}_{2}\in \mathbb{R}^{\mathop{chan}(i)}$ are used to channel-wisely adjust the contributions of $f^{(i)}$ and $g^{(i)}$, where $\mathop{chan}(i)$ denotes the dimension of feature channel. 
To satisfy the assumption of unit-variance activations in the style modulation~\cite{stylegan2}, we add a $\mathop{\mathrm{Conv}}$ layer after each $f^{(i)}$ to initially adjust the activations, and normalize the scaling vectors to be positive and to have channel-wise unit L2 norm.
Formally, we define the feature modulation as follows
\begin{equation}
  \begin{aligned}
    \bm{\sigma} & = \mathop{\mathrm{Reshape}}(\mathop{\mathrm{MLP}}(s)), \\
    \sigma^{(i)}_{enc/gen} & = \frac{|\sigma^{(i)}_{1/2}|}{\sqrt{{\sigma^{(i)}_1}^2 + {\sigma^{(i)}_2}^2 + \epsilon}}, \qquad l\leq i\leq u,\\
    h^{(i)} & = \sigma^{(i)}_{enc} \cdot \mathop{\mathrm{Conv}}(f^{(i)}) + \sigma^{(i)}_{gen} \cdot g^{(i)},
  \end{aligned}
  \label{eqn:feature_mod}
\end{equation}
where $\epsilon = 1\mathrm{e-8}$ and the last two equations are both channel-wise operations.

\paragraph{Output.}
The output image is progressively computed from the fused feature, $h^{(i)}$, via $\mathop{\mathrm{tRGB}}$ layer~\cite{stylegan2}. We up-sample and sum all the intermediate RGB outputs to derive the final output, $\hat{y} = \hat{y}^{(u)}$:
\begin{equation}
  \hat{y}^{(i)} =
  \begin{cases}
    \mathop{\mathrm{tRGB}}(h^{(i)}), & i = l, \\
    \mathop{\uparrow_2}(\hat{y}^{(i-1)}) + \mathop{\mathrm{tRGB}}(h^{(i)}), & l< i\leq u.
  \end{cases}
  \label{eqn:output}
\end{equation}

\subsection{Training Details} 
%
Recall that GCFSR takes the LR image $\bm{x}$ and the conditional upscaling factor $s$ as input and estimates an SR face image, $\hat{\bm{y}} = G(\bm{x}, s)$.
We create the LR images by down-sampling the ground-truth images and then up-sample them to the original size by bicubic interpolation. During training, the upscaling factor $s$ is randomly chosen from $\{4, 8, 16, 32, 64\}$ and normalized to the range of $[0, 1]$. 

GCFSR could be well trained with adversarial loss, where $\bm{y}$ and $\hat{\bm{y}} = G(\bm{x}, s)$ are treated as real and fake images respectively. We use the non-saturating logistic loss: 
\begin{equation}
  \begin{aligned}
    \mathcal{L}_{adv, D} & = \mathbb{E}_{\bm{y}, s} \left[ \log{( 1+\exp{(-D(\bm{y}))} )} \right. \\
    & \left. + \log{(1+\exp{(D(G(\mathop{\downarrow_s}(\bm{y}), s)))} )} \right] \\
    \mathcal{L}_{adv, G} & = \mathbb{E}_{\bm{y}, s} \left[ \log{(1+\exp{(-D(G(\mathop{\downarrow_s}(\bm{y}), s)))} )} \right].
  \end{aligned}
  \label{eqn:loss_adv}
\end{equation}
As shown in Table \ref{table:GCFSR_adv}, our GCFSR$_{adv}$ trained with only adversarial loss outperforms other blind face restoration methods for $4\times$ and $8\times$ SR tasks. 

To further boost the SR performance, we use the conventional combinations: L1, perceptual~\cite{perceptual} and adversarial losses. The overall training objectives are as follows:
\begin{equation}
  \begin{aligned}
    \mathcal{L}_{D} & = \lambda_{adv} \mathcal{L}_{adv, D}, \\
    \mathcal{L}_{G} & = \lambda_{l1}\cdot  \mathcal{L}_1 + \lambda_{perc} \mathcal{L}_{perc} + \lambda_{adv} \mathcal{L}_{adv, G}.
  \end{aligned}
  \label{eqn:loss_total}
\end{equation}
The hyper-parameters are set as: $\lambda_{l1}=1$, $\lambda_{perc}=0.01$, and $\lambda_{adv}=0.01$. 
$D$ and $G$ are trained to minimize $\mathcal{L}_{D}$ and $\mathcal{L}_{G}$ respectively. 
Although GCFSR is trained from scratch, it converges fast (see Figure~\ref{fig:convergence}) and achieves the best performance among the GAN-prior-based methods.

\subsection{Testing}
During testing, the upscaling factor of the given LR image could not be strictly in the set of $\{4, 8, 16, 32, 64\}$. 
A by-product is that GCFSR can achieve continuous SR effect for upscaling factors in the range of $[4, 64]$.
The users are encouraged to modulate the factor $s$ to obtain various super-resolved images with different generative strength and the best one could be found during the modulation process. 
For example, given a LR image downscaled with factor $48$, the users may find the satisfactory results between $s=32$ and $s=64$. As shown in Figure \ref{fig:figure1}, \ref{fig:modulation}, the modulation process yields smooth transitions without artifacts.

\section{Experiments}

\paragraph{Datasets and Implementation}
We train our GCFSR on the FFHQ dataset \cite{ffhq}, consisting of 70k high-quality $1024\times1024$ face images. 
For testing dataset, we follow GLEAN \cite{glean} to extract 100 images from CelebA-HQ \cite{celeb} dataset. We use bicubic interpolation to perform downscaling/upscaling. 
For evaluation, we employ the widely used non-reference perceptual metrics: FID \cite{fid} and NIQE \cite{niqe}. We also adopt pixel-wise metrics (PSNR and SSIM) and the perceptual metric (LPIPS \cite{lpips}). In addition, we measure the cosine similarity on the ArcFace \cite{arcface} embedding space. The training mini-batch size is set to 24. We augment the training data with horizontal flip. We train our model with Adam optimizer for a total of 300k iterations. The learning rates for the generator and discriminator are both set to $2\times10^{-3}$. We implement our models with the PyTorch framework and train them using a NVIDIA Tesla V100 GPU.

\begin{figure*}[htbp]
	\centering
	\includegraphics[scale=0.12]{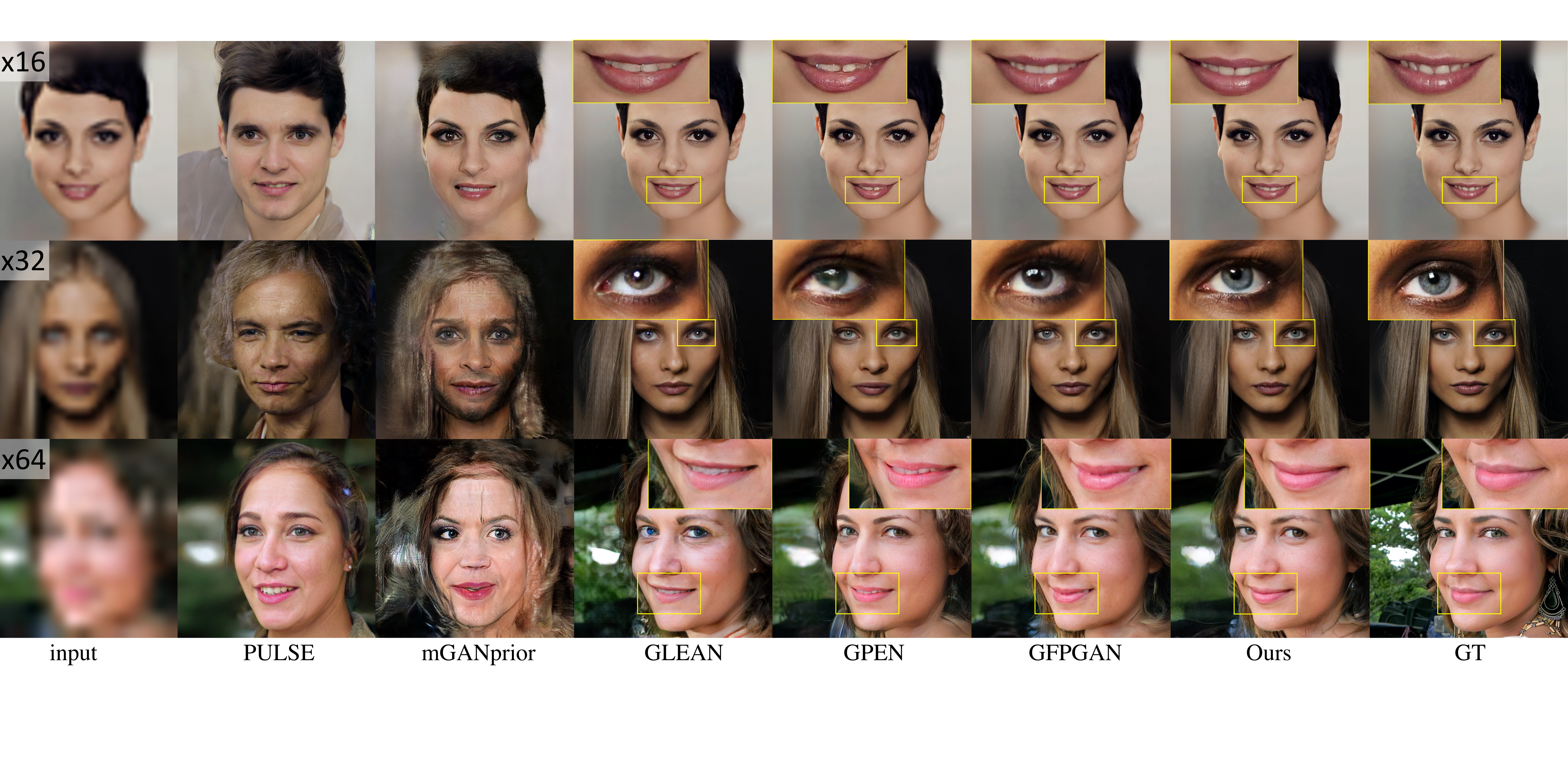}
\vspace{-1em}
	\caption{Qualitative comparisons on CelebA-HQ for $16\times$ (first row), $32\times$ (second row), $64\times$ (third row) SR. The GT image (Right) has a resolution of $1024^2$. \textbf{Zoom in for best view.}}
	\label{fig:main}
\end{figure*}

\begin{table*}[t]
\centering
\scriptsize
\renewcommand{\arraystretch}{1.1}
\setlength{\tabcolsep}{2pt}

\caption{Quantitative comparisons with state-of-the-art methods on CelebA-HQ for $16\times$, $32\times$, $64\times$ SR. GLEAN uses three models, while the others use a single model for three SR tasks. {\color{red}\textbf{Red}} and {\color{blue}blue} indicate the best and the second best performance. Similarity represents cosine similarity of ArcFace Embeddings.}
\vspace{-1em}
\begin{tabular}{@{}l|ccccc|ccccc|ccccc@{}}
\hline
 & \multicolumn{5}{|c|}{$16\times$ ($64^2\rightarrow1024^2$)} & \multicolumn{5}{c|}{$32\times$ ($32^2\rightarrow1024^2$)} & \multicolumn{5}{c}{$64\times$ ($16^2\rightarrow1024^2$)} \\ \hline
 & PSNR $\uparrow$ & SSIM $\uparrow$ & LPIPS $\downarrow$ & FID $\downarrow$ & similarity $\uparrow$ & PSNR $\uparrow$ & SSIM $\uparrow$ & LPIPS $\downarrow$ & FID $\downarrow$ & similarity $\uparrow$ & PSNR $\uparrow$ & SSIM $\uparrow$ & LPIPS $\downarrow$ & FID $\downarrow$ & similarity $\uparrow$\\ \hline
PULSE \cite{pulse} & 20.08 & 0.6032 & 0.4651 & 77.89 & 0.4947 & 19.63 & 0.5848 & 0.4789 & 78.26 & 0.5177 & 18.52 & 0.5604 & 0.5119 & 91.55 & 0.4680 \\
mGANprior \cite{mganprior} & 23.80 & 0.6674 & 0.4668 & 100.87 & 0.6794 & 21.26 & 0.6117 & 0.5099 & 105.62 & 0.5230 & 18.69 & 0.5721 & 0.5530 & 108.17 & 0.4397 \\
GLEAN \cite{glean} & 26.88 & 0.6953 & \color{blue}0.2693 & \color{red}\textbf{29.99} & \color{red}\textbf{0.9682} & 24.34 & 0.6534 & 0.3257 & 46.57 & 0.7750 & 21.38 & 0.6016 & 0.4109 & 62.93 & 0.6118 \\
GPEN \cite{gpen} & 26.51 & 0.6988 & 0.2827 & 37.94 & 0.9473 & 24.65 & 0.6717 & 0.3340 & 50.40 & 0.7641 & 22.20 & 0.6291 & 0.3906 & 67.50 & 0.5978 \\
GFPGAN \cite{gfpgan} & \color{blue}27.07 & \color{red}\textbf{0.7101} & 0.2716 & 34.49 & 0.9623 & \color{blue}24.81 & \color{red}\textbf{0.6751} & \color{blue}0.3128 & \color{blue}46.00 & \color{blue}0.7881 & \color{blue}22.26 & \color{blue}0.6285 & \color{blue}0.3675 & \color{blue}59.33 & \color{blue}0.6558 \\ \hline
\textbf{GCFSR (Ours)} & \color{red}\textbf{27.17} & \color{blue}0.7100 & \color{red}\textbf{0.2604} & \color{blue}30.48 & \color{blue}0.9631 & \color{red}\textbf{24.95} & \color{blue}0.6748 & \color{red}\textbf{0.3061} & \color{red}\textbf{43.34} & \color{red}\textbf{0.7911} & \color{red}\textbf{22.39} & \color{red}\textbf{0.6315} & \color{red}\textbf{0.3663} & \color{red}\textbf{57.15} & \color{red}\textbf{0.6620}\\\hline
\end{tabular}
\label{table:main}
\end{table*}

\paragraph{Comparison with state-of-the-art methods.}
We compare our GCFSR with several state-of-the-art methods: GAN inversion methods including PULSE \cite{pulse} and mGANprior \cite{mganprior}, and GAN-prior-based methods including GLEAN \cite{glean}, GFPGAN \cite{gfpgan} and GPEN \cite{gpen}. 
We provide the quantitative and qualitative results for $16\times$, $32\times$, $64\times$ SR tasks. Note that GLEAN is designed for SR task on single upscaling factor, thus we train three GLEAN models for different tasks.
For fair comparison, we train GFPGAN, GPEN and our proposed GCFSR on the same training dataset towards the same learning objectives. In GFPGAN, the parameters of GAN prior are fixed during training, while the parameters of GAN prior are given a smaller learning rate ($2\times10^{-4}$) in GPEN as suggested by \cite{gpen}. 

The quantitative results are presented in Table \ref{table:main}. As can be seen, our GCFSR achieves the best performance in terms of PSNR and LPIPS for all three upscaling factors, indicating the superiority of GCFSR's ability in image reconstruction with high perceptual quality. 
Although GLEAN is trained on a single upscaling factor, it only performs well on $16\times$ SR. As GLEAN adopts a simple progressive upsampling strategy in the network design without any skip connections, the reconstruction quality cannot be guaranteed when the upscaling factor is large. 
On the other hand, the GAN inversion methods, PULSE and mGANprior, achieve significantly worse results compared with GAN-prior-based methods since they can hardly preserve the identity by the simple latent code exploration strategy.

We also show qualitative results in Figure \ref{fig:main}. It is observed that GAN inversion methods fail to maintain a good fidelity, while the GAN prior based methods achieve overall satisfactory results in term of the identity preservation.
However, as shown in the second row in Figure \ref{fig:main}, GLEAN and GFPGAN are unable to maintain the original color of the eyes since their GAN prior is fixed during training. GPEN performs better than them but shows some distortions. In general, our proposed GCFSR could achieve both fidelity and naturalness with the highest quality among all those methods.
More visual comparisons could be found in the supplementary file.

\begin{figure*}[t]
	\centering
	\includegraphics[scale=0.135]{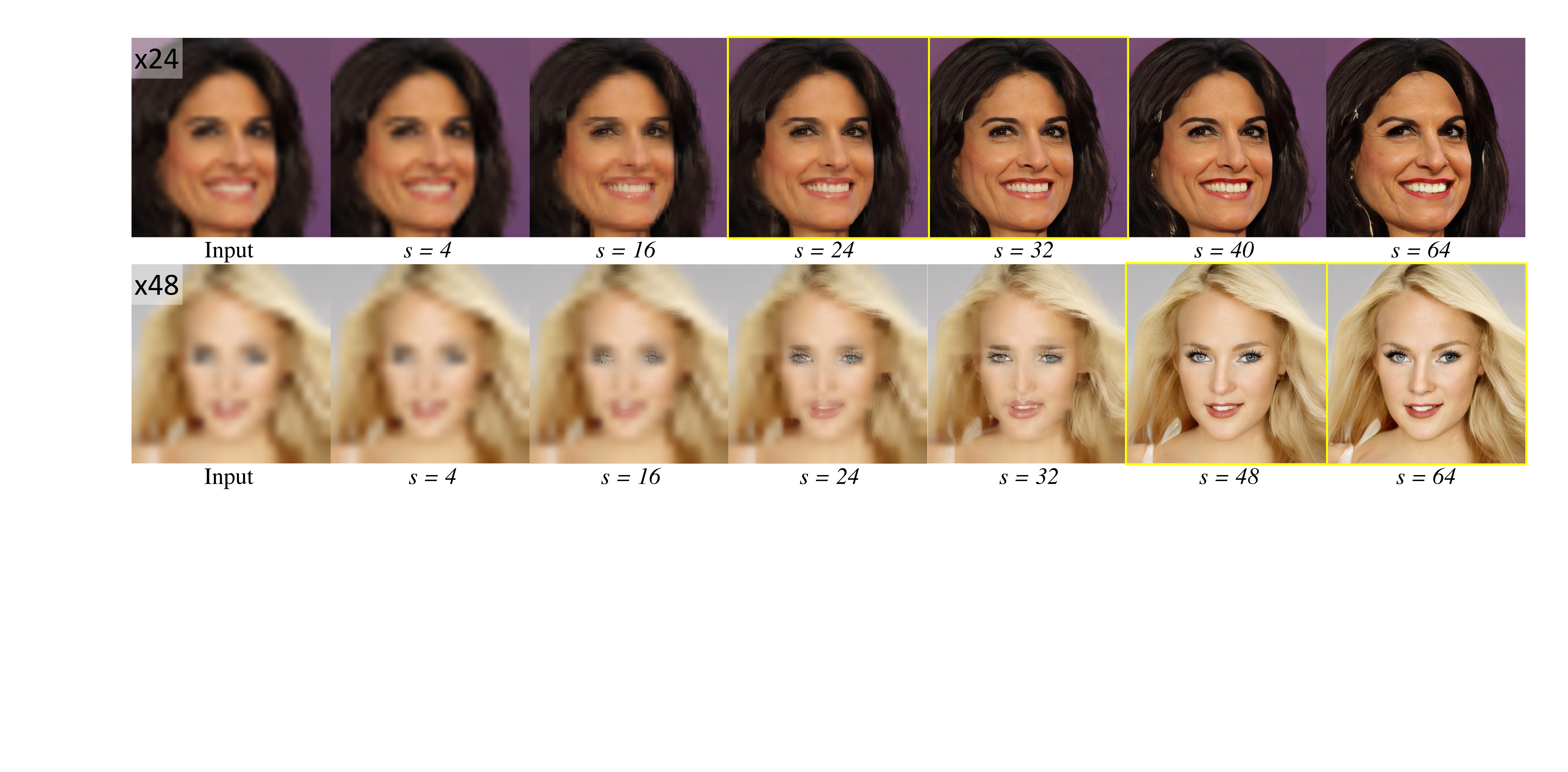}
	\caption{The results obtained by modulation on the generative strength. We change the conditional upscaling factor $s$ from $s=4$ to $s=64$ continuously, and find satisfactory results (e.g., results denoted by yellow rectangles) between two ends. \textbf{Zoom in for best view.}}
	\label{fig:modulation}
\end{figure*}

\paragraph{Evaluation of modulation on generative strength.}
In this section, we provide qualitative results in Figure \ref{fig:figure1}, \ref{fig:modulation} to illustrate that our GCFSR could modulate the generative strength smoothly across different levels. 
In Figure \ref{fig:figure1}, the target upscaling factor $32$ is in the predefined set $\{4, 8, 16, 32, 64\}$, so we could directly specify the conditional upscaling factor to $s=32$ and obtain an SR result with high perceptual quality. Furthermore, we could modulate $s$ around $s=32$ and obtain a blurry effect ($s=24$) or a strong generative effect ($s=48$). 
For target upscaling factor that is not predefined, we could still obtain satisfactory results through modulation.
The LR image in the first row of Figure \ref{fig:modulation} is downscaled with factor $24$. As can be seen, the modulated results between $s=24$ and $s=32$ (denoted by yellow rectangles) are all satisfactory.
Similarly, for LR image downscaled with factor $48$, we could modulate the conditional upscaling factor $s$ continuously within the range of $[48, 64]$ and then arrive at a point with vivid and natural texture details (e.g., $s=48$). The modulation process yields smooth transition without any noticeable artifacts.
More results are in the supplementary file.

\paragraph{The Effectiveness on Blind Face Restoration} In this section, we investigate the effectiveness of our method on blind face restoration (BFR) task. 
To create the blind version of our GCFSR, we fix the upscaling factor $s$ to a constant value (e.g., 1). As for testing dataset, we create CelebA-Test with 3,000 CelebA-HQ images from its testing partition \cite{celeb}.
All the images of FFHQ and CelebA-Test datasets are resized to $512^{2}$. Then, we adopt the degradation model in GFPGAN \cite{gfpgan} to synthesize the training and test input images. Note that we directly use the officially released models of the state-of-the-art blind face restoration methods: DFDNET \cite{dfdnet}, PSFRGAN \cite{psfrgan}, GPEN \cite{gpen}, and GFPGAN \cite{gfpgan}.
The quantitative results are presented in Table \ref{table:blind}. 
It is observed that our blind model could achieve the best performance in PSNR, SSIM, LPIPS, and the cosine similarity of ArcFace Embeddings. Besides, we could obtain comparable results in terms of FID. This indicates the effectiveness of our method for blind face restoration. Please refer to the supplementary file for visual comparison.

\begin{table}[htbp]
\small
\renewcommand{\arraystretch}{1.1}
\setlength{\tabcolsep}{3pt}
\caption{Quantitative comparison on CelebA-Test for blind
face restoration. {\color{red}\textbf{Red}} and {\color{blue}blue} indicate the best and the second best performance. Similarity represents cosine similarity of ArcFace Embeddings.}
\label{table:blind}
\vspace{-1em}
\centering
\begin{tabular}{l|ccccc}
\hline
 & PSNR $\uparrow$ & SSIM $\uparrow$ & LPIPS $\downarrow$ & FID $\downarrow$ & similarity $\uparrow$ \\ \hline
DFDNET & 23.51 & 0.6674 & 0.4342 & 58.72 & 0.5980 \\
PSFRGAN & \color{blue}24.66 & 0.6439 & 0.4199 & 43.33 & 0.6464 \\
GPEN & 24.63 & 0.6477 & 0.4004 & \color{red}\textbf{41.99} & 0.6993 \\
GFPGAN & 24.65 & \color{blue}0.6725 & \color{blue}0.3646 & 42.61 & \color{blue}0.7156 \\
\hline
Ours & \color{red}\textbf{26.49} & \color{red}\textbf{0.7120} & \color{red}\textbf{0.3356} & \color{blue}42.23 & \color{red}\textbf{0.7257} \\ \hline
\end{tabular}
\end{table}

\begin{figure}[htbp]
\begin{minipage}[]{\linewidth}
\centering
\begin{minipage}[]{0.5\linewidth}
	    \centering
		$\sigma^{64}_{enc}$\\
		\includegraphics[scale=0.4]{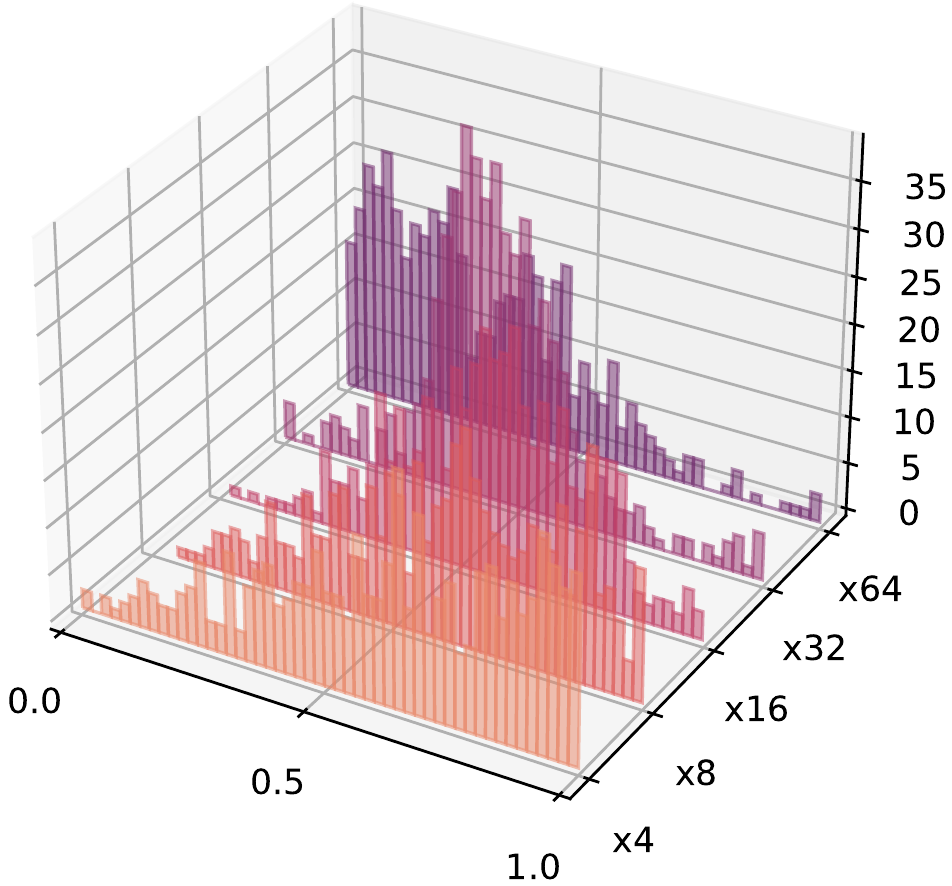}
	\end{minipage}%
	\begin{minipage}[]{0.5\linewidth}
	    \centering
	    $\sigma^{64}_{gen}$\\
		\includegraphics[scale=0.4]{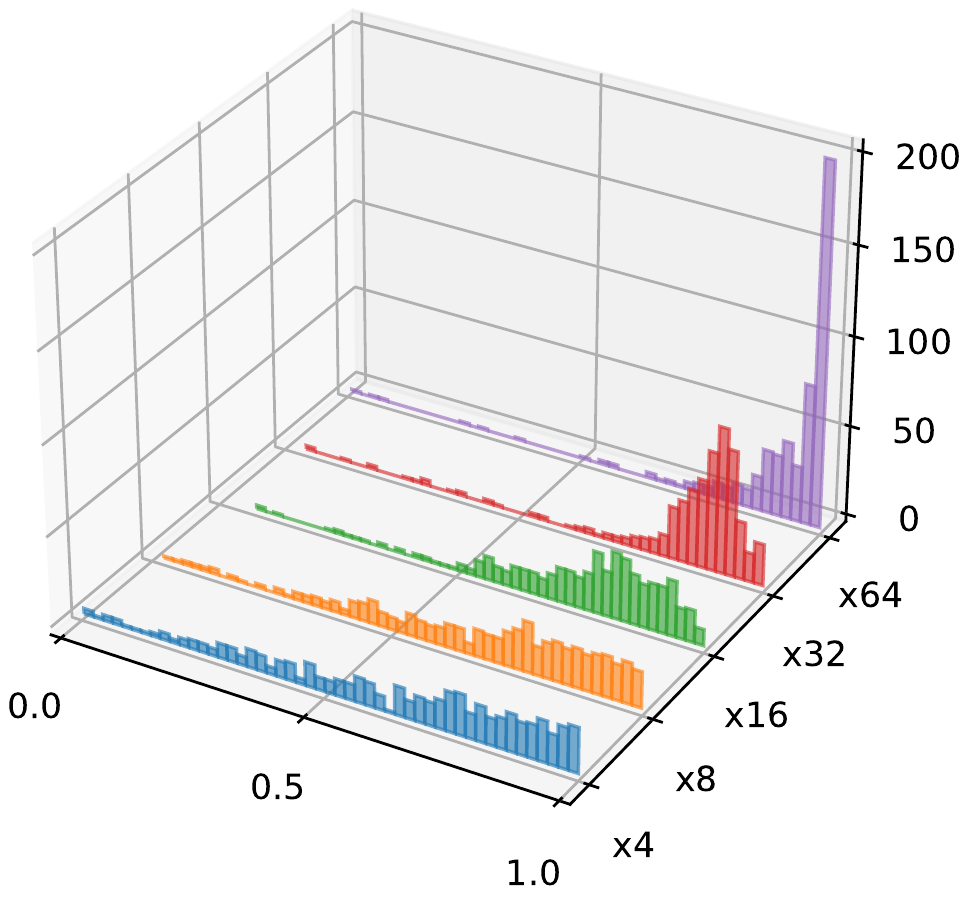}
	\end{minipage}%
\end{minipage}
	\caption{The visualization on feature modulation. The histograms of scaling vectors $\sigma^{64}_{enc}$ and $\sigma^{64}_{gen}$ for different conditional upscaling factors are presented.}
	\label{fig:visual_scale}
\end{figure}

\paragraph{Analysis and visualization on feature modulation.}
In this section, we investigate the mechanism of how the scaling vectors work to achieve effective feature modulation for different conditional upscaling factors. 
As we have mentioned before, the scaling vectors $\bm{\sigma}_{enc}$ and $\bm{\sigma}_{gen}$ channel-wisely adjust the features from the encoder and generator, respectively. Here we provide the histograms of scaling vectors that correspond to level $64$: $\sigma^{64}_{enc}$ and $\sigma^{64}_{gen}$, which are illustrated in Figure \ref{fig:visual_scale}.
For $\sigma^{64}_{enc}$, its values are approaching $0$ as the conditional upscaling factor $s$ increases. Reversely, the values of $\sigma^{64}_{gen}$ are approaching $1$. This indicates that higher conditional upscaling factor corresponds to stronger generative effect, since the features from the encoder are weakened while the features from the decoder are strengthened. 
Similar trend could be found at other levels, presented in the supplementary file.

\begin{table}[htbp]
\centering
\scriptsize
\renewcommand{\arraystretch}{1.1}
\setlength{\tabcolsep}{3pt}
\caption{Quantitative comparison of GCFSR$_{adv}$ (trained with only one adversarial loss) and state-of-the-art blind face restoration methods on CelebA-HQ for $4\times$, $8\times$ SR. {\color{red}\textbf{Red}} and {\color{blue}blue} indicate the best and the second best performance.}
\begin{tabular}{l|ccccc}
\hline  
 & \multicolumn{5}{c}{$4\times$ ($128^2\rightarrow512^2$)} \\\hline
 & PSNR $\uparrow$ & SSIM $\uparrow$ & LPIPS $\downarrow$ & NIQE $\downarrow$ & FID $\downarrow$ \\  \hline
GFPGAN \cite{gfpgan} & 27.32 & 0.7686 & 0.1421 & 4.42 & 36.76 \\
GPEN \cite{gpen} & 27.10 & 0.7593 & 0.1534 & 4.18 & 43.81 \\
HiFaceGAN \cite{hifacegan} & \color{blue}27.39 & 0.7397 & \color{blue}0.1409 & \color{blue}4.06 & \color{blue}30.28 \\
DFDNet \cite{dfdnet} & 26.47 & \color{red}\textbf{0.7802} & 0.1838 & 4.41 & 45.80 \\
PSFRGAN \cite{psfrgan} & 27.24 & 0.7607 & 0.1611 & 4.79 & 35.85 \\\hline
\textbf{GCFSR$_{adv}$ (Ours)} & \color{red}\textbf{27.81} & \color{blue}0.7711 & \color{red}\textbf{0.1210} & \color{red}\textbf{3.84} & \color{red}\textbf{27.90} \\ \hline
 & \multicolumn{5}{c}{$8\times$ ($64^2\rightarrow512^2$)} \\\hline
 & PSNR $\uparrow$ & SSIM $\uparrow$ & LPIPS $\downarrow$ & NIQE $\downarrow$ & FID $\downarrow$ \\  \hline
GFPGAN \cite{gfpgan} & 25.63 & \color{blue}0.7069 & \color{blue}0.1736 & 4.26 & \color{blue}42.88 \\
GPEN \cite{gpen} & \color{blue}25.94 & 0.7038 & 0.1848 & 4.31 & 47.32 \\
HiFaceGAN \cite{hifacegan} & 25.47 & 0.6774 & 0.2294 & \color{red}\textbf{3.61} & 61.52 \\
DFDNet \cite{dfdnet} & 25.03 & 0.7023 & 0.2313 & 4.60 & 52.98 \\
PSFRGAN \cite{psfrgan} & 24.90 & 0.6871 & 0.2085 & 4.58 & 47.10 \\\hline
\textbf{GCFSR$_{adv}$ (Ours)} & \color{red}\textbf{26.02} & \color{red}\textbf{0.7139} & \color{red}\textbf{0.1704} & \color{blue}4.14 & \color{red}\textbf{39.99}\\\hline
\end{tabular}
\label{table:GCFSR_adv}
\end{table}

\paragraph{Effects of learning objectives.}
Our GCFSR is very easy to train and could achieve state-of-the-art performance for large-factor image super resolution. In this section, we will show that our GCFSR could obtain surprisingly good results for small-factor image super resolution ($4\times$, $8\times$) by only using adversarial learning (without pixel-wise loss or perceptual loss). 
Here we compare our GCFSR$_{adv}$ with blind face restoration (BFR) methods: GFPGAN \cite{gfpgan}, GPEN \cite{gpen}, DFDNet \cite{dfdnet}, PSFRGAN \cite{psfrgan}, and HiFaceGAN \cite{hifacegan}.
Note that most of the abovementioned methods only release models for output size $512\times512$. Therefore, for the convenience of comparison, we train our GCFSR$_{adv}$ on resized $512\times512$ FFHQ training dataset and test it on resized $512\times512$ CelebA-HQ dataset. The upscaling factor $s$ is randomly sampled from $\{4, 8, 16, 32\}$. The quantitative results are presented in Table \ref{table:GCFSR_adv}. It observed that our GCFSR$_{adv}$ achieves superior performance to blind face restoration methods. 
We also show visual comparisons among these methods in the supplementary file.


\paragraph{Effects of style modulation.}
Here we demonstrate the effectiveness of the style modulation module. 
As we mentioned before, the latent codes $\bm{w}$ are estimated by the encoder and will be used for style modulation in the generator. Unlike GAN inversion methods which utilize latent codes $\bm{w}$ to generate both global attributes (e.g., poses) and finer details, our $\bm{w}$ is more related to the latter. From Figure \ref{fig:latent}, the results with style modulation have more realistic details (e.g., eyelash and hair) and less artifacts (e.g., bad case for mouth generation). More results are in supplementary file. Besides, we provide the quantitative results in the supplementary file. In general,
the style modulation module improves the overall performance in most metrics.

\begin{figure}[htbp]
	\centering
	\includegraphics[scale=0.36]{latent.pdf}
	\caption{Visual comparison of GCFSR with and without style modulation module on CelebA-HQ for $64\times$ (first row), $32\times$ (second row) and $16\times$ (third row) SR. \textbf{Zoom in for best view.}}
	\label{fig:latent}
\end{figure}



\section{Conclusion}
We have presented a face SR framework called GCFSR without any additional priors but could handle very large-factor face SR (up to $64\times$). GCFSR has an encoder-generator architecture and is end-to-end trainable with fast convergence. In particular, the proposed style modulation module helps generate realistic face details, while the feature modulation module dynamically fuses the multi-level encoded features and generated ones under control of the conditional upscaling factor. In this way, our GCFSR could reconstruct faithful images with promising identity information and provide the flexibility for user adjustment.

\paragraph{Limitations}
Our work has several limitations. First, this work only investigates the single dimensional modulation regarding the upscaling factor. While in real-world scenarios, multi-dimensional modulation across multiple degradations should be considered. 
Second, although our framework shows extraordinary performance on face SR task, its generalization on general SR is waiting to be studied.



\begin{thebibliography}{10}\itemsep=-1pt

\bibitem{biggan}
Andrew Brock, Jeff Donahue, and Karen Simonyan.
\newblock Large scale gan training for high fidelity natural image synthesis.
\newblock In {\em International Conference on Learning Representations}, 2018.

\bibitem{bulat2018super}
Adrian Bulat and Georgios Tzimiropoulos.
\newblock Super-fan: Integrated facial landmark localization and
  super-resolution of real-world low resolution faces in arbitrary poses with
  gans.
\newblock In {\em Proceedings of the IEEE Conference on Computer Vision and
  Pattern Recognition}, pages 109--117, 2018.

\bibitem{cao2017attention}
Qingxing Cao, Liang Lin, Yukai Shi, Xiaodan Liang, and Guanbin Li.
\newblock Attention-aware face hallucination via deep reinforcement learning.
\newblock In {\em Proceedings of the IEEE Conference on Computer Vision and
  Pattern Recognition}, pages 690--698, 2017.

\bibitem{glean}
Kelvin~CK Chan, Xintao Wang, Xiangyu Xu, Jinwei Gu, and Chen~Change Loy.
\newblock Glean: Generative latent bank for large-factor image
  super-resolution.
\newblock In {\em Proceedings of the IEEE/CVF Conference on Computer Vision and
  Pattern Recognition}, pages 14245--14254, 2021.

\bibitem{psfrgan}
Chaofeng Chen, Xiaoming Li, Lingbo Yang, Xianhui Lin, Lei Zhang, and Kwan-Yee~K
  Wong.
\newblock Progressive semantic-aware style transformation for blind face
  restoration.
\newblock In {\em Proceedings of the IEEE/CVF Conference on Computer Vision and
  Pattern Recognition}, pages 11896--11905, 2021.

\bibitem{fsrnet}
Yu Chen, Ying Tai, Xiaoming Liu, Chunhua Shen, and Jian Yang.
\newblock Fsrnet: End-to-end learning face super-resolution with facial priors.
\newblock In {\em Proceedings of the IEEE Conference on Computer Vision and
  Pattern Recognition}, pages 2492--2501, 2018.

\bibitem{arcface}
Jiankang Deng, Jia Guo, Niannan Xue, and Stefanos Zafeiriou.
\newblock Arcface: Additive angular margin loss for deep face recognition.
\newblock In {\em Proceedings of the IEEE/CVF Conference on Computer Vision and
  Pattern Recognition}, pages 4690--4699, 2019.

\bibitem{srcnn}
Chao Dong, Chen~Change Loy, Kaiming He, and Xiaoou Tang.
\newblock Image super-resolution using deep convolutional networks.
\newblock {\em IEEE Transactions on Pattern Analysis \& Machine Intelligence},
  38(02):295--307, 2016.

\bibitem{fsrcnn}
Chao Dong, Chen~Change Loy, and Xiaoou Tang.
\newblock Accelerating the super-resolution convolutional neural network.
\newblock In {\em European conference on computer vision}, pages 391--407.
  Springer, 2016.

\bibitem{swagan}
Rinon Gal, Dana Cohen, Amit Bermano, and Daniel Cohen-Or.
\newblock Swagan: A style-based wavelet-driven generative model.
\newblock {\em arXiv preprint arXiv:2102.06108}, 2021.

\bibitem{mganprior}
Jinjin Gu, Yujun Shen, and Bolei Zhou.
\newblock Image processing using multi-code gan prior.
\newblock In {\em Proceedings of the IEEE/CVF conference on computer vision and
  pattern recognition}, pages 3012--3021, 2020.

\bibitem{fid}
Martin Heusel, Hubert Ramsauer, Thomas Unterthiner, Bernhard Nessler, and Sepp
  Hochreiter.
\newblock Gans trained by a two time-scale update rule converge to a local nash
  equilibrium.
\newblock {\em Advances in neural information processing systems}, 30, 2017.

\bibitem{huang2017wavelet}
Huaibo Huang, Ran He, Zhenan Sun, and Tieniu Tan.
\newblock Wavelet-srnet: A wavelet-based cnn for multi-scale face super
  resolution.
\newblock In {\em Proceedings of the IEEE International Conference on Computer
  Vision}, pages 1689--1697, 2017.

\bibitem{perceptual}
Justin Johnson, Alexandre Alahi, and Li Fei-Fei.
\newblock Perceptual losses for real-time style transfer and super-resolution.
\newblock In {\em European conference on computer vision}, pages 694--711.
  Springer, 2016.

\bibitem{progan}
Tero Karras, Timo Aila, Samuli Laine, and Jaakko Lehtinen.
\newblock Progressive growing of gans for improved quality, stability, and
  variation.
\newblock {\em arXiv preprint arXiv:1710.10196}, 2017.

\bibitem{celeb}
Tero Karras, Timo Aila, Samuli Laine, and Jaakko Lehtinen.
\newblock Progressive growing of gans for improved quality, stability, and
  variation.
\newblock In {\em International Conference on Learning Representations}, 2018.

\bibitem{stylegan3}
Tero Karras, Miika Aittala, Samuli Laine, Erik H{\"a}rk{\"o}nen, Janne
  Hellsten, Jaakko Lehtinen, and Timo Aila.
\newblock Alias-free generative adversarial networks.
\newblock {\em arXiv preprint arXiv:2106.12423}, 2021.

\bibitem{stylegan1}
Tero Karras, Samuli Laine, and Timo Aila.
\newblock A style-based generator architecture for generative adversarial
  networks.
\newblock In {\em Proceedings of the IEEE/CVF Conference on Computer Vision and
  Pattern Recognition}, pages 4401--4410, 2019.

\bibitem{ffhq}
Tero Karras, Samuli Laine, and Timo Aila.
\newblock A style-based generator architecture for generative adversarial
  networks.
\newblock In {\em Proceedings of the IEEE/CVF Conference on Computer Vision and
  Pattern Recognition}, pages 4401--4410, 2019.

\bibitem{stylegan2}
Tero Karras, Samuli Laine, Miika Aittala, Janne Hellsten, Jaakko Lehtinen, and
  Timo Aila.
\newblock Analyzing and improving the image quality of stylegan.
\newblock In {\em Proceedings of the IEEE/CVF Conference on Computer Vision and
  Pattern Recognition}, pages 8110--8119, 2020.

\bibitem{kim2019progressive}
Deokyun Kim, Minseon Kim, Gihyun Kwon, and Dae-Shik Kim.
\newblock Progressive face super-resolution via attention to face landmark.
\newblock In {\em the 30th British Machine Vision Conference (BMVC) 2019}. the
  30th British Machine Vision Conference (BMVC) 2019, 2019.

\bibitem{dfdnet}
Xiaoming Li, Chaofeng Chen, Shangchen Zhou, Xianhui Lin, Wangmeng Zuo, and Lei
  Zhang.
\newblock Blind face restoration via deep multi-scale component dictionaries.
\newblock In {\em European Conference on Computer Vision}, pages 399--415.
  Springer, 2020.

\bibitem{pulse}
Sachit Menon, Alexandru Damian, Shijia Hu, Nikhil Ravi, and Cynthia Rudin.
\newblock Pulse: Self-supervised photo upsampling via latent space exploration
  of generative models.
\newblock In {\em Proceedings of the ieee/cvf conference on computer vision and
  pattern recognition}, pages 2437--2445, 2020.

\bibitem{niqe}
Anish Mittal, Rajiv Soundararajan, and Alan~C Bovik.
\newblock Making a “completely blind” image quality analyzer.
\newblock {\em IEEE Signal processing letters}, 20(3):209--212, 2012.

\bibitem{dgp}
Xingang Pan, Xiaohang Zhan, Bo Dai, Dahua Lin, Chen~Change Loy, and Ping Luo.
\newblock Exploiting deep generative prior for versatile image restoration and
  manipulation.
\newblock {\em IEEE Transactions on Pattern Analysis and Machine Intelligence},
  2021.

\bibitem{psp}
Elad Richardson, Yuval Alaluf, Or Patashnik, Yotam Nitzan, Yaniv Azar, Stav
  Shapiro, and Daniel Cohen-Or.
\newblock Encoding in style: a stylegan encoder for image-to-image translation.
\newblock In {\em Proceedings of the IEEE/CVF Conference on Computer Vision and
  Pattern Recognition}, pages 2287--2296, 2021.

\bibitem{unet}
Olaf Ronneberger, Philipp Fischer, and Thomas Brox.
\newblock U-net: Convolutional networks for biomedical image segmentation.
\newblock In {\em International Conference on Medical image computing and
  computer-assisted intervention}, pages 234--241. Springer, 2015.

\bibitem{shen2018deep}
Ziyi Shen, Wei-Sheng Lai, Tingfa Xu, Jan Kautz, and Ming-Hsuan Yang.
\newblock Deep semantic face deblurring.
\newblock In {\em Proceedings of the IEEE Conference on Computer Vision and
  Pattern Recognition}, pages 8260--8269, 2018.

\bibitem{tuzel2016global}
Oncel Tuzel, Yuichi Taguchi, and John~R Hershey.
\newblock Global-local face upsampling network.
\newblock {\em arXiv preprint arXiv:1603.07235}, 2016.

\bibitem{gfpgan}
Xintao Wang, Yu Li, Honglun Zhang, and Ying Shan.
\newblock Towards real-world blind face restoration with generative facial
  prior.
\newblock In {\em Proceedings of the IEEE/CVF Conference on Computer Vision and
  Pattern Recognition}, pages 9168--9178, 2021.

\bibitem{realesrgan}
Xintao Wang, Liangbin Xie, Chao Dong, and Ying Shan.
\newblock Real-esrgan: Training real-world blind super-resolution with pure
  synthetic data.
\newblock In {\em Proceedings of the IEEE/CVF International Conference on
  Computer Vision}, pages 1905--1914, 2021.

\bibitem{sftgan}
Xintao Wang, Ke Yu, Chao Dong, and Chen~Change Loy.
\newblock Recovering realistic texture in image super-resolution by deep
  spatial feature transform.
\newblock In {\em Proceedings of the IEEE conference on computer vision and
  pattern recognition}, pages 606--615, 2018.

\bibitem{esrgan}
Xintao Wang, Ke Yu, Shixiang Wu, Jinjin Gu, Yihao Liu, Chao Dong, Yu Qiao, and
  Chen Change~Loy.
\newblock Esrgan: Enhanced super-resolution generative adversarial networks.
\newblock In {\em Proceedings of the European conference on computer vision
  (ECCV) workshops}, pages 0--0, 2018.

\bibitem{hifacegan}
Lingbo Yang, Shanshe Wang, Siwei Ma, Wen Gao, Chang Liu, Pan Wang, and Peiran
  Ren.
\newblock Hifacegan: Face renovation via collaborative suppression and
  replenishment.
\newblock In {\em Proceedings of the 28th ACM International Conference on
  Multimedia}, pages 1551--1560, 2020.

\bibitem{gpen}
Tao Yang, Peiran Ren, Xuansong Xie, and Lei Zhang.
\newblock Gan prior embedded network for blind face restoration in the wild.
\newblock In {\em Proceedings of the IEEE/CVF Conference on Computer Vision and
  Pattern Recognition}, pages 672--681, 2021.

\bibitem{yu2018face}
Xin Yu, Basura Fernando, Bernard Ghanem, Fatih Porikli, and Richard Hartley.
\newblock Face super-resolution guided by facial component heatmaps.
\newblock In {\em Proceedings of the European conference on computer vision
  (ECCV)}, pages 217--233, 2018.

\bibitem{yu2016ultra}
Xin Yu and Fatih Porikli.
\newblock Ultra-resolving face images by discriminative generative networks.
\newblock In {\em European conference on computer vision}, pages 318--333.
  Springer, 2016.

\bibitem{bsrgan}
Kai Zhang, Jingyun Liang, Luc Van~Gool, and Radu Timofte.
\newblock Designing a practical degradation model for deep blind image
  super-resolution.
\newblock {\em arXiv preprint arXiv:2103.14006}, 2021.

\bibitem{zhang2018super}
Kaipeng Zhang, Zhanpeng Zhang, Chia-Wen Cheng, Winston~H Hsu, Yu Qiao, Wei Liu,
  and Tong Zhang.
\newblock Super-identity convolutional neural network for face hallucination.
\newblock In {\em Proceedings of the European conference on computer vision
  (ECCV)}, pages 183--198, 2018.

\bibitem{lpips}
Richard Zhang, Phillip Isola, Alexei~A Efros, Eli Shechtman, and Oliver Wang.
\newblock The unreasonable effectiveness of deep features as a perceptual
  metric.
\newblock In {\em Proceedings of the IEEE conference on computer vision and
  pattern recognition}, pages 586--595, 2018.

\bibitem{ranksrgan}
Wenlong Zhang, Yihao Liu, Chao Dong, and Yu Qiao.
\newblock Ranksrgan: Generative adversarial networks with ranker for image
  super-resolution.
\newblock In {\em Proceedings of the IEEE/CVF International Conference on
  Computer Vision}, pages 3096--3105, 2019.

\bibitem{zhujiapeng_inversion}
Jiapeng Zhu, Yujun Shen, Deli Zhao, and Bolei Zhou.
\newblock In-domain gan inversion for real image editing.
\newblock In {\em European conference on computer vision}, pages 592--608.
  Springer, 2020.

\bibitem{zhu2016deep}
Shizhan Zhu, Sifei Liu, Chen~Change Loy, and Xiaoou Tang.
\newblock Deep cascaded bi-network for face hallucination.
\newblock In {\em European conference on computer vision}, pages 614--630.
  Springer, 2016.

\end{thebibliography}

{\small
\bibliographystyle{ieee_fullname}

}

\end{document}